\documentclass[final,5p,times,twocolumn]{elsarticle}

\usepackage{amssymb}
\usepackage{amsmath}
\usepackage{booktabs}
\usepackage{multirow}

\journal{Information Fusion}

\begin{document}

\begin{frontmatter}



\title{Bidirectional Uncertainty-Aware Region Learning for Semi-Supervised Medical Image Segmentation}

\author[1]{Shiwei Zhou}
\ead{wa24101008@stu.ahu.edu.cn}

\author[1]{Xin Liu}
\ead{wa24201024@stu.ahu.edu.cn}

\author[2]{Haifeng Zhao}
\ead{senith@ahu.edu.cn}

\author[2]{Bin Luo}
\ead{luobin@ahu.edu.cn}

\author[1,3]{Dengdi Sun\corref{cor}}
\cortext[cor]{Corresponding author}
\ead{sundengdi@ahu.edu.cn}

\address[1]{Key Laboratory of Intelligent Computing \& Signal Processing (ICSP), Ministry of Education, School of Artificial Intelligence, Anhui Uniersity, Hefei, 230601, China}
\address[2]{Anhui Provincial Key Laboratory of Multimodal Cognitive Computation, School of Computer Science and Technology, Anhui University, Hefei, 230601, China}
\address[3]{Institute of Artificial Intelligence, Hefei Comprehensive National Science Center, Hefei, 230026, China}

\begin{abstract}
In semi-supervised medical image segmentation, the poor quality of unlabeled data and the uncertainty in the model's predictions lead to models that inevitably produce erroneous pseudo-labels. These errors accumulate throughout model training, thereby weakening the model's performance. We found that these erroneous pseudo-labels are typically concentrated in high-uncertainty regions. Traditional methods improve performance by directly discarding pseudo-labels in these regions, which can also result in neglecting potentially valuable training data. To alleviate this problem, we propose a bidirectional uncertainty-aware region learning strategy to fully utilize the precise supervision provided by labeled data and stabilize the training of unlabeled data. Specifically, in the training labeled data, we focus on high-uncertainty regions, using precise label information to guide the model's learning in potentially uncontrollable areas. Meanwhile, in the training of unlabeled data, we concentrate on low-uncertainty regions to reduce the interference of erroneous pseudo-labels on the model. Through this bidirectional learning strategy, the model's overall performance has significantly improved. Extensive experiments show that our proposed method achieves significant performance improvement on different medical image segmentation tasks.
\end{abstract}




\end{frontmatter}

\section{Introduction}
With the continuous development of deep learning, many medical image segmentation tasks~\cite{ronneberger2015u,milletari2016v,shu2022cross} have achieved great success through fully supervised learning using large amounts of labeled data. However, it is challenging to obtain large-scale and accurately labeled medical datasets to train segmentation models due to the privacy, dispersion, and labeling difficulties of medical images~\cite{tajbakhsh2020embracing}. Semi-supervised medical image segmentation~\cite{zhang2024interteach,wang2023mcf,wang2023dual,yao2022enhancing} effectively reduces the cost of labeled data and improves the performance and generalization ability of the model to a certain extent by utilizing a small amount of labeled data and a large amount of easily available unlabeled data to train the model.

\begin{figure}[t]
\includegraphics[width=1.0\columnwidth]{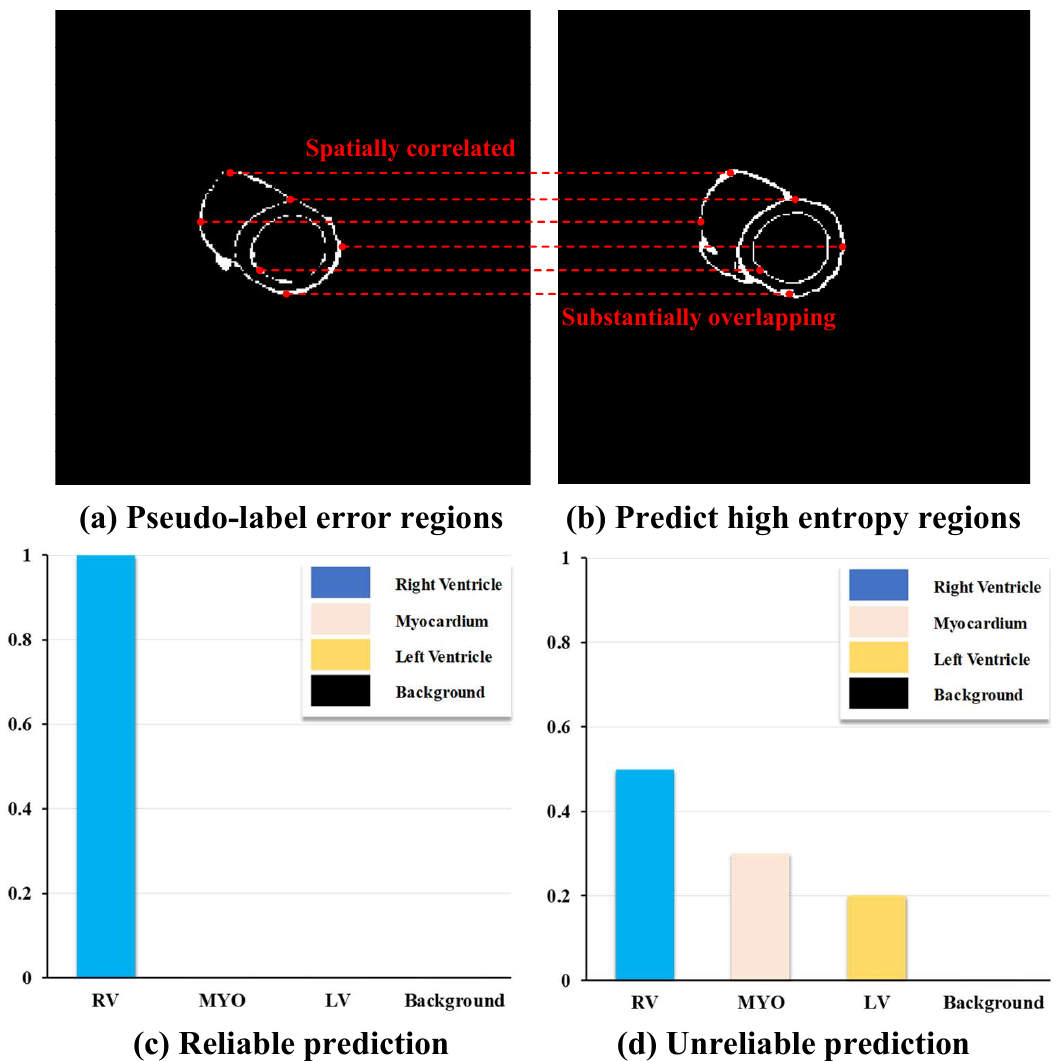}
\centering
\caption{Visualization of (a) pseudo-label error regions, (b) predicted high-entropy regions at the 99th percentile, (c) category probabilities for reliable predictions, and (d) category probabilities for unreliable predictions. The results show that most pseudo-label error regions strongly overlap with the model’s predicted high-entropy regions. Although unreliable predictions fluctuate across multiple categories, they can still effectively exclude certain categories (e.g., in (d) they exclude the background).}
\label{fig:problem}
\end{figure}

In semi-supervised medical image segmentation, effectively leveraging unlabeled data is crucial. A common approach is assigning pseudo-labels to unlabeled data~\cite{basak2023pseudo,he2021re}, allowing the model to learn from these approximate “ground-truth". However, generating high-quality pseudo-labels becomes particularly difficult due to the scarcity of labeled data. Low-quality pseudo-labels introduce greater uncertainty and contain numerous errors, which can mislead the model during training and ultimately affect its performance~\cite{lu2023uncertainty}. Therefore, effectively preventing the accumulation and propagation of errors in pseudo-labels has become a critical challenge in this area of research. Some researchers have proposed predictive filtering based on confidence scores to solve this problem. They argue that pseudo-tags generated by predictions with higher confidence scores are more reliable~\cite{rizve2021defense,zhang2021flexmatch,xu2021dash}. This approach aims to filter out low-confidence predictions and to retrain the model using only high-confidence pseudo-labels. However, the potential issue with relying solely on reliable predictions is that, while incorrect predictions are discarded, the model may never learn certain pixels throughout the training process, resulting in inadequate model training~\cite{wang2022semi}. Particularly in medical images, each region may contain critical medical information~\cite{zhang2022boostmis}. If the model consistently to avoid these challenging regions during training, it will not only make it difficult to learn discriminative features for these regions, but will also cause the model's performance on such regions to deteriorate, thus weakening the model's ability to recognize critical structures. In addition, accurately defining whether an image is of high or low confidence is inherently challenging, further exacerbating this strategy's instability. 

To mitigate the issue of erroneous pseudo-labels more effectively, we analyze the model's predicted entropy maps and the pseudo-label error maps obtained by comparing pseudo-labels with ground-truth labels, as shown in Fig.~\ref{fig:problem} (a) and (b), we find that incorrect pseudo-labels often occur in high-entropy regions, which largely coincide with the boundaries of the segmented targets. Therefore, we can effectively detect and correct pseudo-label errors by identifying and processing these high-entropy regions. Moreover, as shown in Fig.~\ref{fig:problem} (c) and (d), although the model's predictions on the high-entropy regions are unreliable, it is very confident that these pixels do not belong to the background. Therefore, effectively utilizing this information can enhance the model's learning. Based on the above analysis, we naturally identify two key challenges grounded in the characteristics of labeled and unlabeled data: \textbf{(1) For labeled data, how can we leverage accurate ground-truth annotations to guide the correction of uncertain predictions. (2) For unlabeled data, in the absence of supervision, how can we avoid propagating errors while still extracting meaningful semantic cues from high-entropy regions?}

To address the two challenges identified above, we propose a bidirectional region learning strategy tailored for semi-supervised segmentation, which consists of two complementary components: \textbf{Uncertainty-aware Region Learning (URL)} for labeled data and \textbf{Certainty-aware Region Learning (CRL)} for unlabeled data.
\textbf{(1) For labeled data}, to correct predictions in high-uncertainty regions, we utilize the ground-truth to guide learning in these areas. Specifically, for each input image, we compute the prediction entropy map to quantify pixel-wise uncertainty. Based on the entropy values, predictions are categorized into high-entropy (uncertain) and low-entropy (certain) regions. During supervised training, we assign higher learning weights to high-entropy regions, encouraging the model to focus on difficult areas where errors are likely to occur. By leveraging precise supervision in these regions, the model gradually learns to identify and correct potential segmentation errors, thereby improving robustness and generalization.
\textbf{(2) For unlabeled data}, in the absence of ground-truth supervision, we adopt a certainty-aware strategy to mitigate the impact of erroneous pseudo-labels in uncertain regions. Rather than discarding uncertain areas entirely (such as UA-MT calculates losses only in the deterministic region), we reduce their contribution to the loss by assigning lower learning weights during self-supervised learning. This soft suppression prevents noisy pseudo-labels from dominating training while still allowing the model to benefit from potentially informative features in a controlled manner. 
By integrating these two region-aware learning strategies, our approach not only guides the model to effectively learn the discriminative features of uncertain regions using the accurate Ground Truth and migrates them throughout the training process to enhance the model's ability to recognize and model uncertain regions; at the same time, it also achieves robust learning of unlabeled data based on the reduction of noise sensitivity. Extensive experiments demonstrate that our approach achieves substantial performance gains across various baselines.

In summary, the contributions of this work consist of the following aspects:
 
\begin{itemize}
    \item We propose a new bi-directional uncertainty-aware learning strategy based on the difference between labeled and unlabeled data characteristics. This strategy involves learning uncertain regions in labeled data as well as certain regions in unlabeled data.
    \item By focusing on the high-uncertainty regions in labeled data, the model's ability to learn difficult samples can be improved; at the same time, focusing on the certainty regions in unlabeled data can help minimize the interference of erroneous pseudo-labels and stabilize the training.
    \item The proposed method can be easily plug-and-play, which can be embedded into different methods to improve its performance. Extensive experiments validate the feasibility of the method.
\end{itemize}

\section{Related Work}
\subsection{Semi-supervised Learning}
Current semi-supervised learning methods~\cite{van2020survey,li2022adaptive} can generally be divided into two categories: consistency regularization~\cite{laine2016temporal,ouali2020semi} and pseudo-labeling~\cite{lee2013pseudo,qiao2018deep,chen2021semi,ran2024pseudo}. Pseudo-labeling refers to using the model's predictions to generate labels for unlabeled samples, treating these predictions as approximate "ground truth" to expand the training dataset and further train the model. Consistency regularization enforces the model’s outputs to be consistent for the inputs under different perturbations. In addition, some studies~\cite{lu2023mutually,lu2023uncertainty} have combined the methods above and applied them to various tasks, showing superior performance. 

\begin{figure*}[t]
\includegraphics[width=1.0\linewidth]{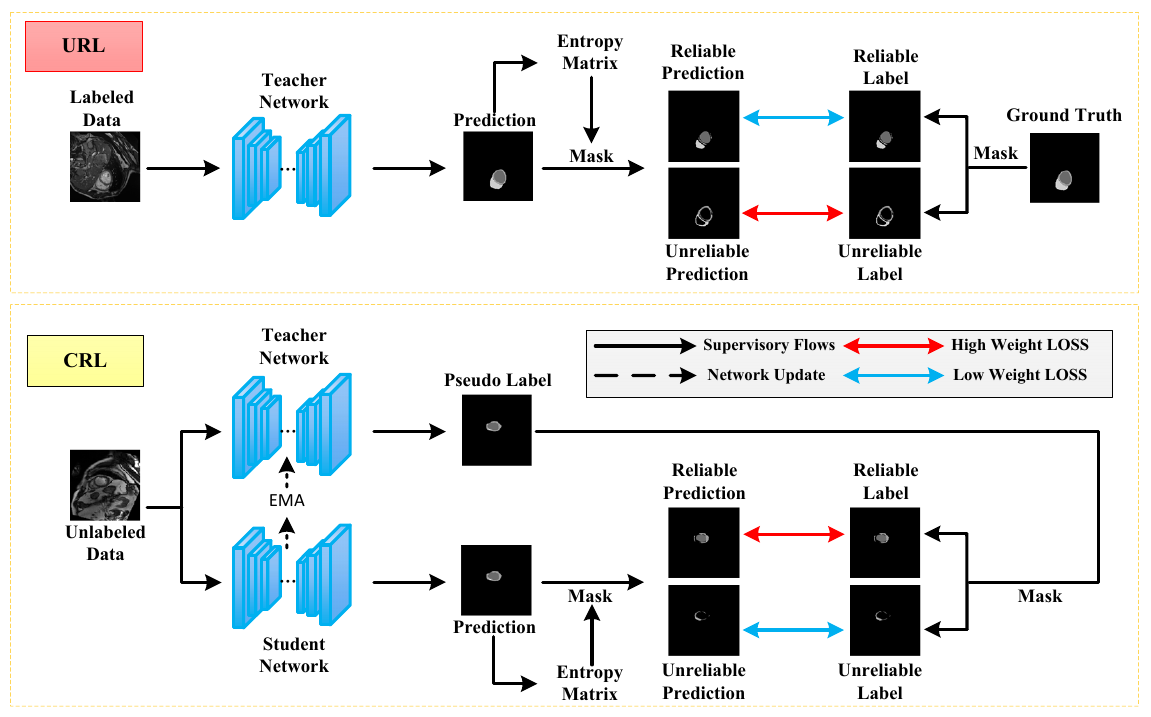}
\centering
\caption{The overall architecture of our proposed bidirectional uncertainty-aware region learning network. Both labeled and unlabeled data stages use entropy-based masks to identify reliable and unreliable predictions. For labeled data, the focus is on learning from uncertain predictions, while for unlabeled data, the focus is on learning from certain predictions.}
\label{fig:framework}
\end{figure*}

\subsection{Semi-Supervised Medical Image Segmentation}
Semantic segmentation is the classification of every pixel in an image~\cite{hao2020brief,mo2022review,cao2022adversarial}. It is a dense prediction task that requires substantial data and meticulous manual annotation in the training phase. To reduce the reliance on large-scale annotated data, many semi-supervised methods have been applied to medical image segmentation. For example, SASSNet~\cite{li2020shape} introduces an edge distance function, calculating the signed distance from any voxel to its nearest natural boundary, and constructing a shape-aware semi-supervised semantic segmentation network. DTC~\cite{luo2021semi} proposes a dual-task consistency framework, introducing dual-task consistency regularization between segmentation maps derived from level sets for labeled and unlabeled data and directly predicted segmentation maps. UA-MT~\cite{yu2019uncertainty}, by incorporating uncertainty-aware consistency loss, guided by the estimated uncertainty, unreliable predictions are filtered out during the calculation of consistency loss, retaining only reliable predictions. BCP~\cite{bai2023bidirectional} alleviates the experience mismatch between labeled and unlabeled data through bidirectional copy-paste operations. ABD~\cite{chi2024adaptive} generates new samples based on adaptive bidirectional patch displacement guided by reliable prediction confidence, followed by training on these new samples to ensure the quality of consistent learning. However, due to the characteristics of medical images, directly removing unreliable regions may lead to the loss of critical anatomical structures, causing the model to overlook important yet difficult-to-detect features, thereby limiting its ability to learn from challenging samples.

\subsection{Uncertainty Estimation}
Uncertainty estimation~\cite{kendall2017uncertainties, mehrtash2020confidence, shi2021inconsistency, cheng2022uncertainty} has been widely explored in semi-supervised learning to calibrate model predictions or enhance the performance of prediction results. For example, Yu and Xia~\cite{yu2019uncertainty,xia2020uncertainty} utilized Monte Carlo Dropout (MC-Dropout)~\cite{gal2016dropout} to estimate uncertainty, which was then employed to guide the model's learning and promote the generation of more accurate pseudo-labels. However, MC-Dropout requires multiple forward passes, leading to significantly increased computational cost and time consumption. Zheng et al.~\cite{zheng2021rectifying} rectified noisy pseudo-labels by estimating uncertainty via the variance across multiple predictions of the same input. Luo et al.~\cite{luo2021efficient} refine the handling of uncertain regions by measuring the deviation between the mean prediction and the scaled prediction. Compared to MC-Dropout, these methods reduce computational overhead. Our method directly estimates uncertainty by calculating the entropy of the model's predictive distribution, a process that does not require additional inference or complex computation and reflects the overall uncertainty stably, mitigating the disturbance of anomalous predictions.

\section{Methodology}
\label{sec:methods}
\subsection{Preliminary}
 In semi-supervised segmentation, we assume that a training dataset $D$ contains $N$ labeled data and $M$ unlabeled data ($N\ll M$). For convenience, we represent them using sets: $D = \mathcal{D}^l \cup \mathcal{D}^u$. Given a small labeled dataset $\mathcal{D}^l = \{(\mathbf{X}^l_i, \mathbf{Y}^l_i)\}_{i=1}^{N}$ and a large number of unlabeled images $\mathcal{D}^u = \{\mathbf{X}^u_i\}_{i=N+1}^{N+M}$, where $\mathbf{X}^l_i$ represents a labeled image, and $\mathbf{Y}^l_i$ represents its corresponding label. Similarly, $\mathbf{X}^u_i$ denotes an unlabeled image. Semi-supervised semantic segmentation aims at mining the unlabeled images with the help of limited labeling under effective supervision to obtain segmentation performance comparable to the results of the same type of segmentation with full supervision. 
 
 \subsection{Overview}
The overall pipeline of the proposed bidirectional uncertainty-aware region learning method is shown in Fig.~\ref{fig:framework}, based on the teacher-student model \cite{tarvainen2017mean} with a teacher network ${F}_t(\theta_t)$ and a student network ${F}_s(\theta_s)$. Due to the differing characteristics of labeled and unlabeled data, our approach to identifying and utilizing uncertainty regions also differs. Specifically, for labeled images, we first train using the teacher network. To enhance the model's ability to learn potentially uncontrollable regions, we use the proposed URL method, which generates a mask by calculating the entropy matrix of the model prediction and using it to divide both the model predictions and the corresponding ground truth into reliable and unreliable regions. We then assign higher learning weights to the unreliable regions, ensuring that the model focuses on uncertain areas during training. For unlabeled data, we use student networks for training and employ pseudo-labels generated by teacher networks trained on labeled data as supervision. To reduce the negative impact of potentially uncontrollable regions on the model, we apply the same mask treatment to the model predictions and pseudo-labels, and use the proposed CRL method to reduce the learning weights of unreliable regions, thus directing the model to mainly learn predictions from deterministic regions, improving the stability of training and the accuracy of the final segmentation.

 \subsection{Uncertainty-aware Region Learning }
In semi-supervised medical image segmentation, regions that are difficult for the model to predict accurately inevitably exist due to factors such as blurry boundaries, similar tissue structures, or poor imaging quality.  To allow the model to learn from potentially uncontrollable regions while maximizing the use of accurate annotation information, we assign higher learning weights to these regions during the training process with labeled data, guiding the model to focus on learning their feature distributions to enhance its understanding and prediction capability in complex regions.

Specifically, we first train the teacher model using labeled data and utilize the entropy matrix to distinguish between reliable and unreliable prediction regions in the labeled data. The entropy matrix reflects the model's uncertainty in classifying each region. If the entropy value is high, it indicates that the prediction for the region is ambiguous, with similar probabilities across categories, making it difficult to determine the region's category. Conversely, a low entropy value suggests that the model's prediction for the region is highly confident, with the probability distribution concentrated on a specific category. Next, we leverage the precise information from the labels to enhance the learning intensity of the unreliable prediction regions, thereby improving the model's prediction ability in these areas.

Suppose $\mathbf{X}^{l}_i$ is a labeled medical image, and after going through the segmentation model, we generate its corresponding outputs:
\begin{equation}
    logits^{l}_i = F_{t}(\mathbf{X}^{l}_i),
\end{equation}
where $F_{t}$ is the teacher network that we need to pre-train. $logits^{l}_i$ are the logits outputs that correspond to $\mathbf{X}^{l}_i$. Applying softmax to these logits yields the corresponding prediction probability scores:
\begin{equation}
    \mathbf{P}^{l}_i = softmax(logits^{l}_i),
\end{equation}
where $\mathbf{P}^{l}_i$ is the prediction of $\mathbf{X}^{l}_i$. With the inference of prediction, uncertainty (i.e., entropy) maps corresponding to each class are computed according to the following equation:
\begin{equation}
    U = p \log p
\end{equation}
where $p$ is the output probability map of each class, that is, for the $i$-th labeled medical image, its value is $\mathbf{P}^{l}_i$. To detect and learn the uncertain regions, we set a dynamic threshold, and if the entropy value of a region exceeds this threshold, the region is considered part of the highly uncertain region. Because labeled data contains accurate labels that can be directly utilized by the model for supervised learning, the model's uncertain regions on labeled data can also be corrected by the label information. Therefore, a lower threshold can be set to include these regions more broadly, thus utilizing the label information more comprehensively in training. After finalizing the uncertainty region, we generate the corresponding image mask based on it $\mathbf{M} (\mathbf{M}\in\{0,1\}^{W\times H\times L})$. The formula is as follows:
\begin{equation}
    \begin{aligned}
&\mu = \mathrm{Percentile}({U}, q),\\
&\mathbf{M} = \left\{
\begin{array}{ll}
1, & \text{if} \; U > \mu \\
0, & \text{otherwise},
\end{array}
\right.
    \end{aligned}
\end{equation}
where $percentile(U,q)$ denotes the function that finds the $q$-th percentile in the predicted probability distribution $U$ of the model each time. We then adjust the labeling of the labeled images utilizing the mask $\mathbf{M}$, The formula is as follows:
\begin{equation}
 \begin{aligned}
    &\mathbf{Y}^{l}_c = \mathbf{Y}^{l}\odot\left( \mathbf{1} - \mathbf{M} \right),\\ 
    &\mathbf{Y}^l_{uc} = \mathbf{Y}^{l}\odot\mathbf{M},
\end{aligned}
\end{equation}
where $\mathbf{Y}^{l}_c$ and $\mathbf{Y}^l_{uc}$ denote, respectively, reliable region labeling and unreliable region labeling.

\subsection{Certainty-aware Region Learning }
Under various perturbations, the segmentation model produces unreliable predictions for unlabeled data~\cite{chi2024adaptive}, and using these predictions may mislead the model to learn the wrong information or features, affecting the learning process and effectiveness of the model. A direct solution is to remove the regions associated with unreliable predictions. However, for medical image segmentation, this approach may result in the loss of important training data. Additionally, the model would not have the opportunity to learn how to handle these complex regions, which are often the most critical and challenging parts of a medical image. We believe that a better approach is to learn these unreliable regions appropriately by adapting the learning strategy, thus preserving valuable information while reducing its adverse effect on model training.

Specifically, for an unlabeled data $\mathbf{X}^{l}_i$, we use the teacher model ${F}_s(\theta_s)$ trained from the labeled data to generate initial pseudo-labels $\mathbf{Y}^{u}$. Similar to the URL, we perform the same operation on the unlabeled data to obtain the uncertainty region and certainty region predicted by the model, and at the same time, generate the uncertainty mask $\mathbf{M}^{\prime}$ according to the entropy threshold set in advance, and classify the pseudo-labels into reliable pseudo-labels and unreliable pseudo-labels by the uncertainty mask $\mathbf{M}^{\prime}$, we would like to reduce the learning intensity of the uncertainty regions in the CRL. It should be noted that after the training of the URL, the prediction results of the model are gradually stabilized, and the uncertain regions show a tendency of concentration and convergence, so we set a higher threshold in the CRL to accurately locate the uncertain regions. If the threshold is too low, some regions accurately predicted by the model may be incorrectly labeled as uncertain regions, thus affecting the training effect. The following formula processes the final pseudo-labeling:
\begin{equation}
 \begin{aligned}
    &\mathbf{Y}^{u}_c = \mathbf{Y}^{u}\odot\left( \mathbf{1} - \mathbf{M^{\prime}} \right),\\
    &\mathbf{Y}^u_{uc} = \mathbf{Y}^{u}\odot\mathbf{M^{\prime}},
\end{aligned}
\end{equation}
where $\mathbf{Y}^{u}_c$ and $\mathbf{Y}^u_{uc}$ denote reliable region pseudo-label and unreliable region pseudo-label, respectively.

\subsection{Loss Functions}
The overall loss function consists of two parts, the supervised loss for labeled data and the unsupervised loss for unlabeled data, defined as: 
\begin{align}
\mathcal{L}^{l} &= {L}_{seg}\left(\mathbf{P}^{l},\mathbf{Y}^{l}_{uc}\right)\odot\mathbf{M} + \alpha{L}_{seg}\left(\mathbf{P}^{l},\mathbf{Y}^{l}_{c}\right)\odot\left(\mathbf{1}-\mathbf{M}\right), \\
\mathcal{L}^{u} &= {L}_{seg}\left(\mathbf{P}^{u},\mathbf{Y}^{u}_{c}\right)\odot\left(\mathbf{1}-\mathbf{M^{\prime}}\right) + \alpha{L}_{seg}\left(\mathbf{P}^{u},\mathbf{Y}^{u}_{uc}\right)\odot\mathbf{M^{\prime}},
\end{align}
where $L_{seg}$ is a hybrid function to compute the Dice loss and Cross-entropy loss, $\mathbf{P}^{l}$ and $\mathbf{P}^{u}$ are the prediction results of the teacher and student model for the labeled data and unlabeled data, respectively. The total loss consists of the loss of labeled and unlabeled images:
\begin{equation}
\mathcal {L}_{all} = \mathcal{L}^{l} + \mathcal{L}^{u},
\end{equation}
In each iteration, we update the parameters in the student network by stochastic gradient descent of the loss function, and the parameters of the teacher network by exponential moving average ($EMA$)~\cite{laine2016temporal}:
\begin{equation}
\theta_t = (1-\lambda) \cdot \theta_{t-1} + \lambda \cdot \theta_s,
\end{equation}
where $t$ and ${t-1}$ denote the current training step and its previous one, respectively, $\theta_{s}$ and $\theta_t$  are the weights of the student and teacher models at training step t. where $\lambda$ is the smoothing coefficient parameter, with a recommended value of $\min\left(t / (t+1), 0.99\right)$, which is used to control the updating rate of the teacher weight.

\section{Experiments and Results}
\subsection{Dataset}
\subsubsection{ACDC dataset}
The automatic Cardiac Diagnosis Challenge (ACDC) dataset comprises scans from 100 patients, categorized into four classes(i.e. background, right ventricle, left ventricle, and myocardium.) The dataset is divided into a fixed allocation of 70 patients' scans for training, 10 for validation, and 20 for testing. 

\subsubsection{LA Dataset}
The Left Atrium (LA) MR dataset from the Atrial Segmentation Challenge dataset includes 100 3D gadolinium-enhanced magnetic resonance image scans (GE-MRIs) with labels, categorized into two classes(i.e. background and left atrium.)

\subsubsection{PROMISE12 Dataset}
The PROMISE12 dataset~\cite{litjens2014evaluation} was made available for the MICCAI 2012 prostate segmentation challenge. Magnetic Resonance (MR) images (T2-weighted) of 50 patients with various diseases were acquired at different locations with several MRI vendors and scanning protocols. All 3D scans are converted into 2D slices for segmentation tasks.

\subsection{Evaluation Metrics}
We choose four evaluation metrics: {Dice Score} (\%), {Jaccard Score} (\%), {95\% Hausdorff Distance (95HD) in voxel} and {Average Surface Distance (ASD) in voxel}. Dice~\cite{ma2021loss} and Jaccard measure overlap percentages between two object regions, while ASD calculates the average boundary distance, and 95HD computes the closest point distance between them. The following are the formulas for these evaluation metrics:
\begin{equation}
\begin{aligned}
&\text{Dice} =  \frac{2|A \cap B|}{|A| + |B|} \times 100\% ,\\
&\text{Jaccard} =  \frac{|A \cap B|}{|A \cup B|} \times 100\% ,\\
&\text{95HD} = \max\{ \sup_{a \in A} \inf_{b \in B} |a - b|, 
 \sup_{b \in B} \inf_{a \in A} |b - a|\},\\
&\text{ASD} = \frac{\sum_{a \in A} \inf_{b \in B} |a - b| + \sum_{b \in B} \inf_{a \in A} |b - a| }{|A| + |B|} .
\end{aligned}
\end{equation}

In these formulas, $A$ and $B$ represent the predicted and true label regions of the image, respectively. The terms $\sup_{a \in A} \inf_{b \in B} |a - b|$ and $\sup_{b \in B} \inf_{a \in A} |b - a|$ denote the maximum of the minimum distances from any point in set $A$ to the closest point in set $B$ and vice versa.
By using these evaluation metrics, we can comprehensively assess the model's performance across different dimensions, including overlap regions and boundary distances, allowing for a more accurate evaluation of the model's effectiveness.

\begin{table}[h!]  
\centering
\setlength{\tabcolsep}{0.6mm}
\resizebox{1\linewidth}{!}{
\begin{tabular}{ccccccc}
\toprule
\textbf{Method} &\textbf{Labeled} &\textbf{Unlabeled} & \textbf{Dice$\uparrow$} & \textbf{Jaccard$\uparrow$} & \textbf{95HD$\downarrow$} & \textbf{ASD$\downarrow$} \\ \midrule
\multicolumn{1}{c}{U-Net} &\multicolumn{1}{c}{5\%} &\multicolumn{1}{c}{0\%} &47.83 &37.01 &31.16 &12.62 \\ 
\multicolumn{1}{c}{U-Net} &\multicolumn{1}{c}{10\%} &\multicolumn{1}{c}{0\%} &79.41 &68.11 &9.35 &2.70 \\
\multicolumn{1}{c}{U-Net} &\multicolumn{1}{c}{100\%} &\multicolumn{1}{c}{0\%} &91.44 &84.59 &4.30 &0.99 \\\midrule
MT & \multirow{12}{*}{5\%} & \multirow{12}{*}{95\%} 
& 51.20 & 40.06 & 21.27 & 6.92 \\
SASSNet &  &  & 57.77 & 46.14 & 20.05 & 6.06 \\ 
DTC & & & 56.90 & 45.67 & 23.36 & 7.39 \\
URPC & & & 55.87 & 44.64 & 13.60 & 3.74 \\ 
MC-Net & & & 62.85 & 52.29 & 7.62 & 2.33  \\
SS-Net & & & 65.83 & 55.38 & 6.67 & 2.28 \\ 
BCP & & & 87.59 & 78.67 & 1.90 & 0.67 \\
GA-BAP & & &  88.24 & 79.60 & 3.91 & 1.11 \\ 
ABD & & & 88.96 & 80.70 & \textbf{1.57} & \textbf{0.52} \\ \midrule
Ours(MT) & & & {63.64}&{51.46}& {20.32}& {6.57} \\ 
Ours(BCP) & & & 88.42&79.83& 1.82& 0.64\\
Ours(ABD) & & & \textbf{89.73}&\textbf{81.92}& 1.86& 0.58\\
\midrule
MT & \multirow{12}{*}{10\%} & \multirow{12}{*}{90\%}& 79.65 & 68.06 & 12.87 & 3.99\\
SASSNet & & & 84.50 & 74.34 & 5.42 & 1.86 \\ 
DTC & & & 84.29 & 73.92 & 12.81 & 4.01  \\
URPC & & & 83.10 & 72.41 & 4.84 & 1.53  \\ 
MC-Net & & & 86.44 & 77.04 & 5.50 & 1.84   \\
SS-Net & & & 86.78 & 77.67 & 6.07 & 1.40 \\
BCP & & & 88.84 & 80.62 & 3.98 & 1.17 \\
GA-BCP& & & 89.31 & 81.27 & 3.32 & 1.01\\ 
ABD & & & 89.81 & 81.95 & \textbf{1.46} & 0.49 \\ \midrule
Ours(MT) & & & {82.19}&{71.36}&{10.04}&{2.66}\\
Ours(BCP) & & & \textbf{90.40}&\textbf{83.01}& 1.63& \textbf{0.41}\\
Ours(ABD) & & & 90.18&82.72& 2.27& 0.69\\
\bottomrule
\end{tabular}}
\caption{Comparisons with state-of-the-art semi-supervised segmentation methods on the ACDC dataset.}
\label{tab:acdc}
\end{table}

\subsection{Implementation Details}
We selected several baseline models for our experiments: Mean Teacher (MT)~\cite{tarvainen2017mean}, BCP~\cite{bai2023bidirectional}, and ABD~\cite{chi2024adaptive}. MT is a basic teacher-student network. BCP builds upon this by introducing a bidirectional copy-paste strategy for labeled and unlabeled data to generate mixed input data. ABD designs an adaptive bidirectional displacement strategy based on prediction confidence to generate new samples for retraining. To verify the effectiveness of our training strategy, we apply our method to these approaches to explore its adaptability to different baseline models and assess the performance improvement it brings. For ABD, we used only one teacher-student network for training in our experiments.

In the experiment, we set the default value of the learning weight to $\alpha = 0.5$. We performed all experiments on an NVIDIA 3090 GPU device with a fixed random seed. Due to the differences between 2D and 3D medical image datasets, our experimental settings are also adjusted accordingly. Specifically, for the 2D ACDC and PROMISE12 datasets, the input size is cropped to $256 \times 256$, we set the threshold $\mu$ to the 95th percentile of the entropy matrix predicted by the model each time during the supervision phase and the threshold $\mu$ to the 99th percentile during the self-training phase. For the 3D LA dataset, the input size is cropped to $112 \times 112 \times 80 $, the threshold $\mu$ was set to the 95th percentile of the entropy matrix in both cases. All other settings follow the default settings with the original method. 

\begin{table}[h!]  
\centering
\setlength{\tabcolsep}{0.7mm}
\resizebox{1\linewidth}{!}{
\begin{tabular}{ccccccc}
\toprule
\textbf{Method} &\textbf{Labeled} &\textbf{Unlabeled} & \textbf{Dice$\uparrow$} & \textbf{Jaccard$\uparrow$} & \textbf{95HD$\downarrow$} & \textbf{ASD$\downarrow$} \\ \midrule
\multicolumn{1}{c}{V-Net} &\multicolumn{1}{c}{5\%} &\multicolumn{1}{c}{0\%} &52.55 &39.60 &47.05 &9.87 \\ 
\multicolumn{1}{c}{V-Net} &\multicolumn{1}{c}{10\%} &\multicolumn{1}{c}{0\%} &82.74 &71.72 &13.35 &3.26 \\
\multicolumn{1}{c}{V-Net} &\multicolumn{1}{c}{100\%} &\multicolumn{1}{c}{0\%} &91.47 &84.36 &5.48 &1.51 \\\midrule
MT & \multirow{10}{*}{5\%} & \multirow{10}{*}{95\%} 
& 81.74 & 69.93 & 16.30 & 4.99 \\
SASSNet &  &  & 81.60 & 69.63 & 16.16 & 3.58 \\ 
DTC & & & 81.25 & 69.33 & 14.90 & 3.99 \\
URPC & & & 82.48 & 71.35 & 14.65 & 3.65 \\ 
MC-Net & & & 83.59 & 72.36 & 14.07 & 2.70  \\
SS-Net & & & 86.33 & 76.15 & 9.97 & 2.31 \\ 
BCP & & & 88.02 & 78.72 & 7.90 & 2.15 \\  
AllSpark& & & 87.99 & 78.83 & 7.44 & 2.10 \\ \midrule
Ours(MT) & & & {82.88}&{71.40}&{11.11}&{3.18}\\
Ours(BCP) & & & \textbf{89.10}&\textbf{80.46}& \textbf{7.72}& \textbf{1.98}\\\midrule
MT & \multirow{10}{*}{10\%} & \multirow{10}{*}{90\%} 
& 86.81 & 76.93 & 10.07 & 2.49 \\
SASSNet & & & 87.54 & 78.05 & 9.84 & 2.59 \\ 
DTC & & & 87.51 & 78.17 & 8.23 & 2.36  \\
URPC & & & 86.92 & 77.03 & 11.13 & 2.28  \\ 
MC-Net & & & 87.62 & 78.25 & 10.03 & 1.82   \\
SS-Net & & & 88.55 & 79.62 & 7.49 & 1.90 \\
BCP & & & 89.62 & 81.31 & 6.81 & 1.76 \\ 
AllSpark & & & 88.74 & 80.54 & 7.06 & 1.82\\ \midrule
Ours(MT) & & & {87.95}&{78.77}&{8.54}&{2.11}\\
Ours(BCP) & & & \textbf{90.21}&\textbf{82.26}& \textbf{6.27}& \textbf{1.63}\\
\bottomrule
\end{tabular}}
\caption{Comparisons with state-of-the-art semi-supervised segmentation methods on LA dataset.}
\label{tab:LA}
\end{table}

\subsection{Comparison with Sate-of-the-Art Methods}
\subsubsection{ACDC dataset}
To assess the generalization capability of our model across different domains, we conducted additional comparative experiments on the ACDC dataset with labeling ratios of 5\% and 10\%.. We present a comparative analysis of our framework against various competing methods: MT~\cite{tarvainen2017mean}, SASSNet~\cite{li2020shape}, DTC~\cite{luo2021semi}, URPC~\cite{luo2021efficient}, MC-Net~\cite{wu2021semi}, SS-Net~\cite{wu2022exploring}, BCP~\cite{bai2023bidirectional}, GA-BCP~\cite{qi2024gradient}, and ABD~\cite{chi2024adaptive}. As shown in Table~\ref{tab:acdc}, incorporating our methods into MT, BCP, and ABD (i.e., Ours(MT), Ours(BCP), and Ours(ABD), they show significant improvements, which highlights the flexibility and scalability of our methods. Specifically, at a labeling data of 5\%, applying our method to the basic MT model (i.e., Ours(MT)) results in performance surpassing several comparison methods. When integrated with BCP and ABD, both baseline models show stable improvements at 5\% and 10\% labeled data. Notably, Ours(ABD) achieves the best performance with a Dice score of 89.73\% at 5\% labeled data. At a 10\% labeling rate, Ours(BCP) reaches a Dice score of 90.40\%, achieving the best result. These results demonstrate that our bidirectional uncertainty-aware region learning strategy effectively enhances the model's focus on difficult region learning while reducing the influence of erroneous pseudo-labels. This dual strategy improves segmentation quality and pushes the upper bound of model performance in semi-supervised settings.

\begin{figure*}[h]
\includegraphics[width=1\linewidth]{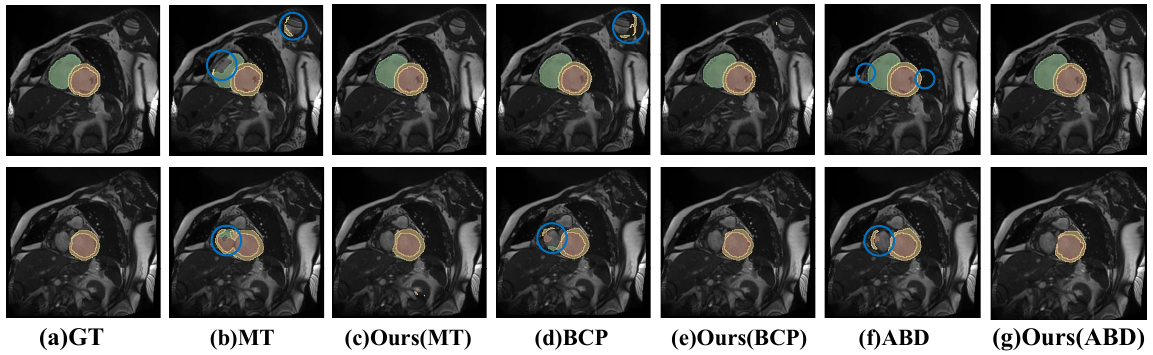}
\centering
\caption{Visualization of segmentation results on ACDC dataset with 10\% labeled data. (a) Ground-truth. (b) Mean Teacher results. (c) Ours (Mean Teacher) results. (d) BCP results. (e) Ours (BCP) results. (f) ABD results. (g) Ours (ABD) results. }
\label{fig:acdc_visio}
\end{figure*}

Fig. \ref{fig:acdc_visio} shows a qualitative comparison of segmentation results on the ACDC dataset, as can be observed from the figure, the baseline methods (MT, BCP, and ABD) are prone to segmentation errors in regions with complex structures or fuzzy boundaries, which are manifested as unclear boundaries or incomplete region prediction. In contrast, our method performs more robustly in these difficult regions and is able to effectively suppress erroneous predictions, presenting clearer structural boundaries and more complete region coverage.

\begin{figure*}[h]
\includegraphics[width=1\linewidth]{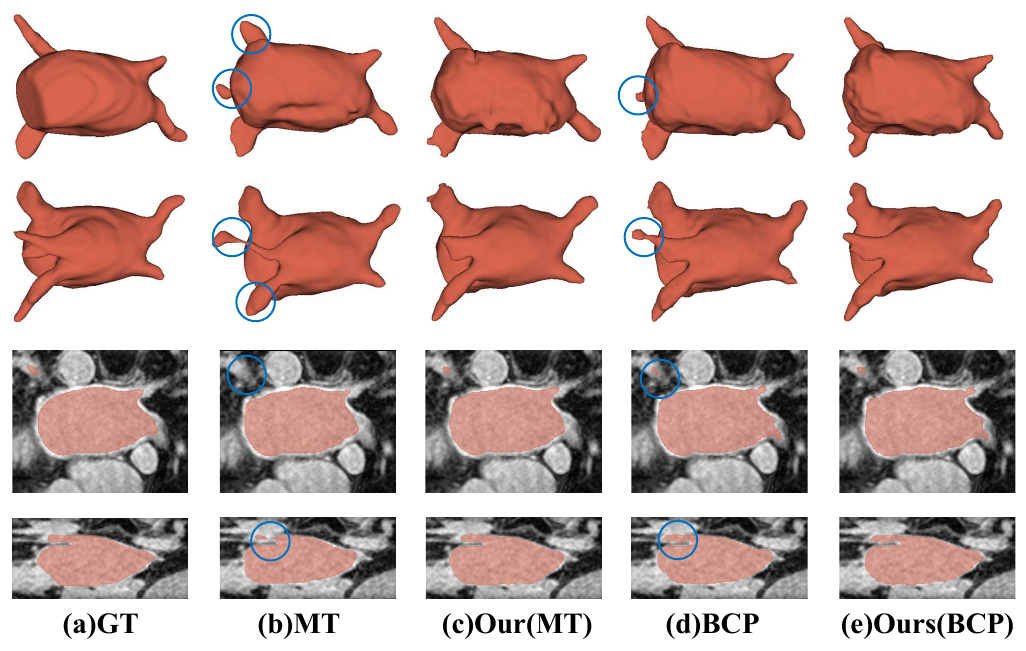}
\centering
\caption{Visualization of segmentation results on LA dataset with 10\% labeled data. (a) Ground-truth. (b) Mean Teacher results. (c) Ours (Mean Teacher) results. (d) BCP results. (e) Ours (BCP) results.}
\label{fig:la_visio}
\end{figure*}

\subsubsection{LA dataset}
Following previous methods, we conducted experiments on the LA dataset using 5\% and 10\% labeled data. We compared our method with several state-of-the-art semi-supervised segmentation approaches, including MT, SASSNet~\cite{li2020shape}, DTC~\cite{luo2021semi}, URPC~\cite{luo2021efficient}, MC-Net~\cite{wu2021semi}, SS-Net~\cite{wu2022exploring}, BCP~\cite{bai2023bidirectional}, and AllSpark~\cite{wang2024allspark}. As shown in Table~\ref{tab:LA}, integrating our method into MT and BCP, i.e., Ours(MT) and Ours(BCP), they both achieves significant improvements, with Ours(BCP) showing the most advanced performance, demonstrating that by applying different uncertainty handling strategies to labeled and unlabeled data, it can also effectively help existing models improve their performance on 3D datasets.

We further performed visualization analysis on the LA dataset to verify the advantages of the proposed strategy in the actual segmentation results. As shown in Fig. \ref{fig:la_visio}, we demonstrate the segmentation results of each method from 3D view and 2D slice level, respectively. It can be observed that the baseline model is prone to prediction errors when dealing with the edges of anatomical structures or regions with complex shapes, while the segmentation performance of the model in these regions is significantly improved after integrating our proposed strategy into the baseline model.

\begin{table}[h!]
\centering
\resizebox{1\linewidth}{!}{
\begin{tabular}{ccccc}
\toprule
\textbf{Method} &\textbf{Labeled} &\textbf{Unlabeled} & \textbf{Dice$\uparrow$} & \textbf{ASD$\downarrow$} \\ \midrule
\multicolumn{1}{c}{U-Net} &\multicolumn{1}{c}{20\%} &\multicolumn{1}{c}{0\%} &60.88 &13.87 \\
\multicolumn{1}{c}{U-Net} &\multicolumn{1}{c}{100\%} &\multicolumn{1}{c}{0\%} &91.47 &84.36 \\\midrule
CCT  &\multirow{7}{*}{10\%} & \multirow{7}{*}{90\%} &46.73  &2.14  \\
URPC &  &  & 61.92 &2.94   \\ 
SS-Net & & & 57.41 &6.33  \\ 
SCP-Net & & &55.14  &19.52  \\ 
BCP  & & &79.68  &7.84 \\ 
ABD & & & 81.81 & 1.46\\ \midrule
Ours(BCP) & & & \textbf{85.20} &\textbf{1.00} \\ 
\midrule
CCT & \multirow{7}{*}{20\%} & \multirow{7}{*}{80\%} & 71.43 & 16.61 \\
URPC &  &  & 63.23 & 4.33  \\ 
SS-Net & & & 62.31& 4.36 \\ 
SCP-Net & & & 77.06 & 3.52 \\ 
BCP  & & &79.81  & 2.9\\ 
ABD & & & 82.06 & 1.33 \\  \midrule
Ours(BCP) & & &  \textbf{86.01} & \textbf{1.04} \\ 
\bottomrule
\end{tabular}}
\caption{Comparisons with state-of-the-art semi-supervised segmentation methods on PROMISE12 dataset.}
\label{tab:PROMISE12}
\end{table}

\subsubsection{PROMISE12 dataset.} 
We also perform the comparison experiments in the PROMISE12 dataset using 10\% and 20\% labeled data. As shown in the Table~\ref{tab:PROMISE12}, We compare our method with CCT~\cite{ouali2020semi}, URPC~\cite{luo2021efficient}, SS-Net~\cite{wu2022exploring}, SCP-Net~\cite{zhang2023self}, BCP~\cite{bai2023bidirectional} and ABD~\cite{chi2024adaptive}. When our strategy is integrated into the BCP framework, the proposed uncertainty-aware region learning method brings consistent and significant performance improvements.  Specifically, under the 10\% labeled setting, Ours(BCP) achieves the best Dice score of 85.20\% and the lowest ASD of 1.00\%, outperforming all other competing methods.  Under the 20\% labeled setting, Ours(BCP) again leads with a Dice score of 86.01\% and an ASD of 1.04\%.  These results further confirm that our method can effectively enhance the performance of strong baseline models and push their upper performance bounds.

\subsection{Ablation Studies}
We conduct an ablation study to show the impact of each component in our model, including the effectiveness of the bidirectional region learning strategy, the regional learning approach, the uncertainty threshold $\mu$, the Weight in the Loss Function $\alpha$, and the impact of removing unreliable prediction regions on model learning.

\begin{table}[h!]
\centering
\setlength{\tabcolsep}{0.7mm}
\resizebox{1\linewidth}{!}{
\begin{tabular}{ccccccc}
\toprule
\textbf{Base} &\textbf{URL} &\textbf{CRL} &\textbf{Dice$\uparrow$}  & \textbf{Jaccard$\uparrow$}  & \textbf{95HD$\downarrow$} &\textbf{ASD$\downarrow$}\\
\midrule
\checkmark &            &   &88.84 & 80.62 & 3.98 & 1.17\\
\checkmark &\checkmark  &    & 89.45 & 81.44 & 6.37 & 1.66 \\
\checkmark & & \checkmark  & 90.20 & 82.69 & 1.83 & 0.43 \\
\checkmark &\checkmark & \checkmark  &\textbf{90.40}&\textbf{83.01}& \textbf{1.63}& \textbf{0.41} \\
\bottomrule
\end{tabular}}
\caption{Effectiveness of URL and CRL modules. “Base” means the baseline is BCP. }
\label{tab:module}
\end{table}

\subsubsection{Effectiveness of Each Module in Bidirectional Region Learning}
As shown in Table~\ref{tab:module}, to validate the bidirectional region learning strategies, we explore the following approaches under Ours(BCP) 10\% labeled data. We demonstrate the effectiveness of each module in dual-phase region learning by progressively adding URL and CRL. When URL is added, the Dice increased to 89.45\%, indicating that focusing on high-uncertainty regions in labeled data helps the model better learn from complex areas and transfer the learned knowledge to the training of unlabeled data. When CRL is added, the Dice rose to 90.20\%, indicating that focusing on low-uncertainty regions in unlabeled data while reducing the learning weights of high-uncertainty regions helps mitigate the impact of erroneous pseudo-labels. By incorporating both URL and CRL into the baseline, the model has the best result, reaching 90.40\% Dice and surpassing the baseline in 1.56\%. These results indicate that the two modules are complementary during the training process. In this way, the model not only gains a deeper understanding of complex regions but also maintains stability and accuracy in the training of unlabeled data, thereby improving the overall performance of the model.

\begin{table}[h!]
\centering
\setlength{\tabcolsep}{0.7mm}
\resizebox{1\linewidth}{!}{
\begin{tabular}{cccccc}
\toprule
\textbf{Labeled} & \textbf{Unlabeled} & \textbf{Dice$\uparrow$} & \textbf{Jaccard$\uparrow$} &\textbf{95HD$\downarrow$} & \textbf{ASD$\downarrow$} \\\midrule
CRL  & CRL   & 90.13 & 82.53 & 2.07 & 0.55 \\
CRL  & URL  & 90.28 & 82.80 & 1.88 & 0.76 \\
URL & URL & 90.12 & 82.58 & 2.04 & 0.54 \\
URL & CRL  &\textbf{90.40}&\textbf{83.01}& \textbf{1.63}& \textbf{0.41} \\
\bottomrule
\end{tabular}}
\caption{Impact of different regional learning strategies on model performance.}
\label{tab:region}
\end{table}

\subsubsection{Regional Learning Strategies}
To evaluate the effectiveness of the URL and CRL modules on labeled and unlabeled data, under 10\% labeled data based on Ours (BCP), we conduct different combination experiments with both modules on labeled and unlabeled data, as shown in Table~\ref{tab:region}. Through these combination strategies, we are able to analyze the effects of applying the two modules on different types of data. The experimental results indicate that all combination strategies brought significant performance improvements, demonstrating the effectiveness of the URL and CRL modules. The optimal result is achieved by applying URL on labeled data and CRL on unlabeled data. This result confirms our hypothesis: for labeled data, we should focus on learning uncertain regions. By concentrating on these uncertain areas and leveraging accurate label information for supervision, the model can gain a deeper understanding of the features in challenging regions, thus improving its segmentation ability in difficult areas. In contrast, for unlabeled data, due to the absence of ground-truth, errors in the pseudo-labels generated by the model are inevitable. Therefore, focusing on learning the deterministic regions predicted by the model can effectively reduce the interference of pseudo-label noise, thus enhancing the stability of the training. The combination of these two strategies fully exploits the different characteristics of labeled and unlabeled data, improving the model's segmentation performance.

\begin{figure}[t]
\includegraphics[width=1.0\columnwidth]{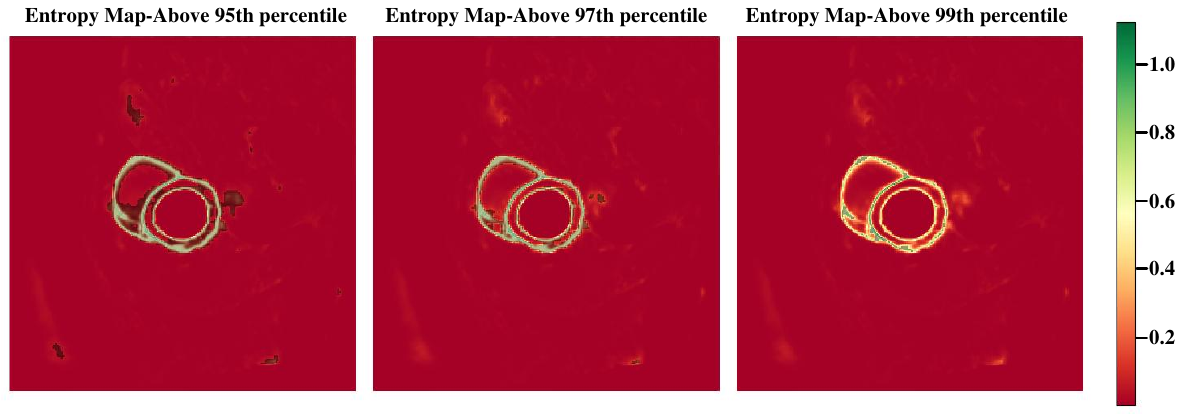}
\centering
\caption{The entropy visualization of the model predictions shows the entropy distribution under three different percentile thresholds (95\%, 97\%, and 99\%). In the images, red represents low-entropy regions, while green indicates high-entropy regions. As the percentile threshold increases, the retained high-entropy regions gradually decrease and are concentrated around the edges of the target regions.}
\label{fig:threshold}
\end{figure}

\begin{table}[t]
\centering
\resizebox{1\linewidth}{!}{
\begin{tabular}{cc|cccc}
\toprule
\textbf{URL} & \textbf{CRL} & \textbf{Dice$\uparrow$} & \textbf{Jaccard$\uparrow$}& \textbf{95HD$\downarrow$} & \textbf{SD$\downarrow$}\\
\midrule
95\% & 95\%   & 90.10 & 82.54 & 3.01 & 0.78 \\
95\% & 99\%  &\textbf{90.40}&\textbf{83.01}& \textbf{1.6}& \textbf{0.41} \\
99\% & 99\% & 90.06 & 82.48 & 2.05 & 0.64 \\
\bottomrule
\end{tabular}}
\caption{Impact of uncertainty threshold $\mu$ settings in URL and CRL on model performance.}
\label{tab:approach4}
\end{table}

\subsubsection{Effect of Uncertainty Thresholds $\mu$ on Model Performance}
In this section, we explore the effect of the uncertainty threshold $\mu$ on model performance. Specifically, choosing the 95th percentile to identify uncertainty regions in labeled data aims to comprehensively cover areas where errors may occur and rely on accurate labels to correct them. Conversely, selecting the 99th percentile in unlabeled data aims to more accurately pinpoint uncertainty regions. As shown in Fig.~\ref{fig:threshold}, when the $\mu$ is set at the 95th and 97th percentiles, some uncertainty regions partially appear in the background. Generally speaking, the model can receive many background samples during training, resulting in a lower probability of prediction errors in background regions. Therefore, the choice of the 99th percentile aims to avoid misclassifying correct regions as uncertainty regions while maximizing the precision of uncertainty region localization. To validate this hypothesis, we conduct the relevant experiments, as shown in Table~\ref{tab:approach4}: Compared to the baseline, the combinations of $\mu$ set at 95\%+95\% and 99\%+99\% for labeled and unlabeled data both exhibit reliable performance, indirectly validating the effectiveness of the proposed module. However, their performance is still lacking compared to the 95\%+99\% combination. This suggests that the mixed approach of using a broader uncertainty range (95\%) in labeled data and a more precise uncertainty selection (99\%) in unlabeled data achieves the optimal balance. This combination effectively provides ample correction opportunities for labeled data while avoiding misjudgment of uncertain regions in unlabeled data. 

\begin{figure}[h]
\includegraphics[width=1.0\columnwidth]{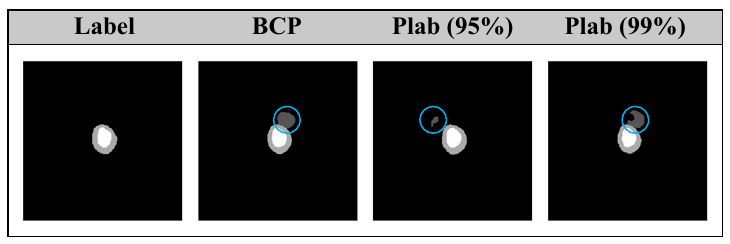}
\centering
\caption{Visualization of pseudo-labels generated for unlabeled data in the unsupervised initial phase using teacher models trained with different threshold values in the supervised phase.}
\label{fig:plab}
\end{figure}

Additionally, in the initial stage of training with unlabeled data, we visualize the pseudo-labels generated by the teacher model. As shown in Fig.~\ref{fig:plab}, in BCP, the pseudo-labels generated by the teacher model contain many errors. However, in Ours(BCP), the pseudo-label errors generated by the teacher model trained with the 99th percentile threshold are slightly reduced. In contrast, those generated with the 95th percentile threshold are significantly fewer. This not only verifies the feasibility of our experimental setup, but also further demonstrates that focusing on learning uncontrollable regions in labeled data can enhance the model's ability to understand and adapt to the uncontrollable regions, thus generating higher-quality pseudo-labels during the training process on unlabeled data, and improving the model's self-supervised learning ability and overall performance.

\begin{table}[h!]   
\centering
\small
\resizebox{1\linewidth}{!}{
\begin{tabular}{ccccc}
\toprule
$\mathbf{\alpha}$ &\textbf{Dice$\uparrow$} & \textbf{Jaccard$\uparrow$} & \textbf{95HD$\downarrow$} & \textbf{ASD$\downarrow$}\\ \midrule
0.2    & 90.05 & 82.36 & 2.25 & 0.62 \\ 
0.5   &\textbf{90.40}&\textbf{83.01}& \textbf{1.63}& \textbf{0.41}  \\
0.7  &90.26 &82.76 & 2.28& 0.58  \\
\bottomrule
\end{tabular}}
\caption{Impact of weighting factor $\alpha$ on model learning.}
\label{tab:approach5}
\end{table}

\subsubsection{Weight in Loss Function}
Table~\ref{tab:approach5} investigates the impact of the weighting factor $\alpha$ on the loss function, evaluated on the ACDC dataset with 10\% labeled data. The results show that setting $\alpha$ = 0.5 achieves the best balance, the configuration used in our experiments. Furthermore, different $\alpha$ settings outperform the original BCP baseline, indicating that our uncertainty region learning strategy remains effective even when the balance between uncertainty and certainty region learning is not optimal. 

\begin{table}[h!]
\centering
\small
\resizebox{1\linewidth}{!}{
\begin{tabular}{ccccc}
\toprule
\textbf{Method} &\textbf{Dice$\uparrow$} & \textbf{Jaccard$\uparrow$} & \textbf{95HD$\downarrow$} & \textbf{ASD$\downarrow$} \\
\midrule
remove   & 88.56 & 80.12 & 4.89 & 1.39 \\
do not remove &\textbf{90.40}&\textbf{83.01}& \textbf{1.63}& \textbf{0.41}  \\\bottomrule
\end{tabular}} 
\caption{Impact of removing unreliable prediction region on model learning.}
\label{tab:approach3} 
\end{table}

\subsubsection{Remove Unreliable Predictions on Model Learning}
 We further investigate the effect of directly removing unreliable predictions on model performance when training the model on the ACDC dataset with 10\% labeled data. As shown in Table~\ref{tab:approach3}, the experimental results show that simply removing these unreliable predictions instead leads to a decrease in model performance. This suggests that even incorrectly predicted regions can still provide valuable training information for the model to some extent. Completely discarding them not only reduces the amount of available training data but may also hinder the model’s ability to adapt to complex or ambiguous regions. In contrast, our method retains these regions and adjusts their influence through adaptive weighting, allowing the model to benefit from difficult samples while minimizing the risk of being misled by noisy supervision. This strategy strikes a better balance between leveraging information and ensuring training stability.
\begin{table}[t]
\centering
\small
\resizebox{1\linewidth}{!}{
\begin{tabular}{ccccc}
\toprule
\textbf{Dataset} & \multicolumn{2}{c}{\textbf{ACDC}} & \multicolumn{2}{c}{\textbf{LA}} \\\midrule
Method & BCP& Ours & BCP &Ours\\ \midrule
FLOPs   & 3.49G & 3.53G & 6.00G & 6.16G \\
Runtime & 3.15 h & 3.52 h & 9.67 h & 10.25 h \\
\bottomrule
\end{tabular}}
\caption{Comparison of training cost between BCP and our method.}
\label{tab:complex}
\end{table}

\subsubsection{Runtime and Efficiency} We compare the FLOPs and runtime between BCP and our method. As shown in Table~\ref{tab:complex}, our method introduces only a slight increase time, primarily attributed to the lightweight entropy-based computation and region-weighting strategy. These additional operations incur minimal computational overhead and do not substantially affect overall training efficiency, demonstrating the practical feasibility of our approach for real-world clinical deployment.

\section{Limitations and Future Directions}
Although our proposed bi-directional uncertainty-aware region learning strategy has demonstrated significant performance advantages in several medical image segmentation tasks, especially in improving the model's ability to perceive difficult regions and attenuating pseudo-labelling misrepresentation, there are still some potential limitations of the method that deserve further exploration. Firstly, the division of uncertainty regions relies on fixed percentile thresholds (e.g., 95\% and 99\%, which may need to be readjusted to suit their characteristics when facing different datasets or more complex tasks. Second, the learning weight adjustment strategy we impose on uncertain regions during training is still an empirical setting. That is, enhancing the learning intensity of high uncertainty regions in labelled data and weakening the learning intensity of uncertainty regions in unlabelled data, this ‘raise-lower’ mechanism is intuitive and effective, but lacks the ability of more fine-grained dynamic adjustment. Therefore, future research can further explore more adaptive mechanisms to improve model stability and generalisation across datasets.

\section{Conclusion}
In this paper, we propose a semi-supervised training method based on a two-way uncertainty-aware region learning strategy, which employs different uncertainty region learning strategies for labeled and unlabeled data, to enhance the model's ability to learn the samples from the difficult regions, and at the same time, reduce the negative impact of these samples on the training of the model. Specifically, for the labeled data, we focus on processing the uncertainty regions to improve the model's segmentation accuracy of the uncertainty regions through precise labeling information. Second, for the unlabeled data, since pseudo-labeling errors cannot be avoided, we adjust the learning focus of the model so that it focuses on learning the deterministic region, which can mitigate the learning interferences introduced by erroneous pseudo-labeling. With the collaboration of these two strategies, the model can better leverage the characteristics of both labeled and unlabeled data, achieving more stable and precise segmentation performance. 

\section*{Declarations}
\noindent\textbf{Funding} This study is funded by the National Natural Science Foundation of China (62076005, U20A20398), the Natural Science Foundation of Anhui Province (2008085MF191), the University Synergy Innovation Program of Anhui Province, China (GXXT-2021-002), and the Provincial Quality Project of Education in the New Era in 2023 (Postgraduate Education 2023lhpysfjd009).

\noindent\textbf{Data Availability} The data that support the findings of this study are available from the corresponding author upon reasonable request.

\noindent\textbf{Confict of Interest} The authors declare that they have no conflict of interest.

\bibliographystyle{unsrt}
\bibliography{cas-refs}

\end{document}